\documentclass[10pt,twocolumn,letterpaper]{article}

\usepackage{wacv}

\usepackage{times}
\usepackage{epsfig}
\usepackage{graphicx}               
\usepackage{amsmath}
\usepackage{amssymb}
\usepackage{subfigure}
\usepackage{multirow}
\usepackage[algoruled,vlined,linesnumbered]{algorithm2e}
\usepackage{rotating}
\usepackage{bbm}
\usepackage{bm}
\usepackage{color}
\usepackage{url}
\usepackage{threeparttable}
\usepackage{multirow}
\usepackage{rotating}
\usepackage{makecell}

\wacvfinalcopy
\usepackage{authblk}

\author{Qi Dong}
\author{Shaogang Gong}
\author{Xiatian Zhu}
\affil{School of EECS, Queen Mary University of London, UK}
\affil{\small\texttt{\{q.dong, s.gong, xiatian.zhu\}@qmul.ac.uk}}

\def\wacvPaperID{141} 

\ifwacvfinal\fi
\begin{document}

\title{Multi-Task Curriculum Transfer Deep Learning of Clothing Attributes}


\maketitle
\ifwacvfinal\thispagestyle{empty}\fi

\begin{abstract}
Recognising detailed clothing characteristics (fine-grained attributes)
in unconstrained images of people in-the-wild is a challenging task
for computer vision, especially when there is only limited training data
from the wild whilst most data available for model learning are
captured in well-controlled environments using fashion models (well
lit, no background clutter, frontal view, high-resolution). In this
work, we develop a deep learning framework capable of model transfer
learning from well-controlled shop clothing images collected from web
retailers to in-the-wild images from the street.  Specifically, we
formulate a novel Multi-Task Curriculum Transfer (MTCT) deep learning
method to explore multiple sources of different types of web
annotations with multi-labelled fine-grained attributes.
Our multi-task loss function is designed to extract more discriminative representations
in training by jointly learning all attributes,
and our curriculum strategy exploits
the staged easy-to-hard transfer learning motivated by cognitive
studies. We demonstrate the advantages of the MTCT model over the
state-of-the-art methods on the X-Domain benchmark, a large scale
clothing attribute dataset.
Moreover, we show that the MTCT model has a notable advantage over contemporary models
when the training data size is small.

\end{abstract}

\section{Introduction}
\label{sec:Intro}
\begin{figure}
	\includegraphics[width=1\linewidth]{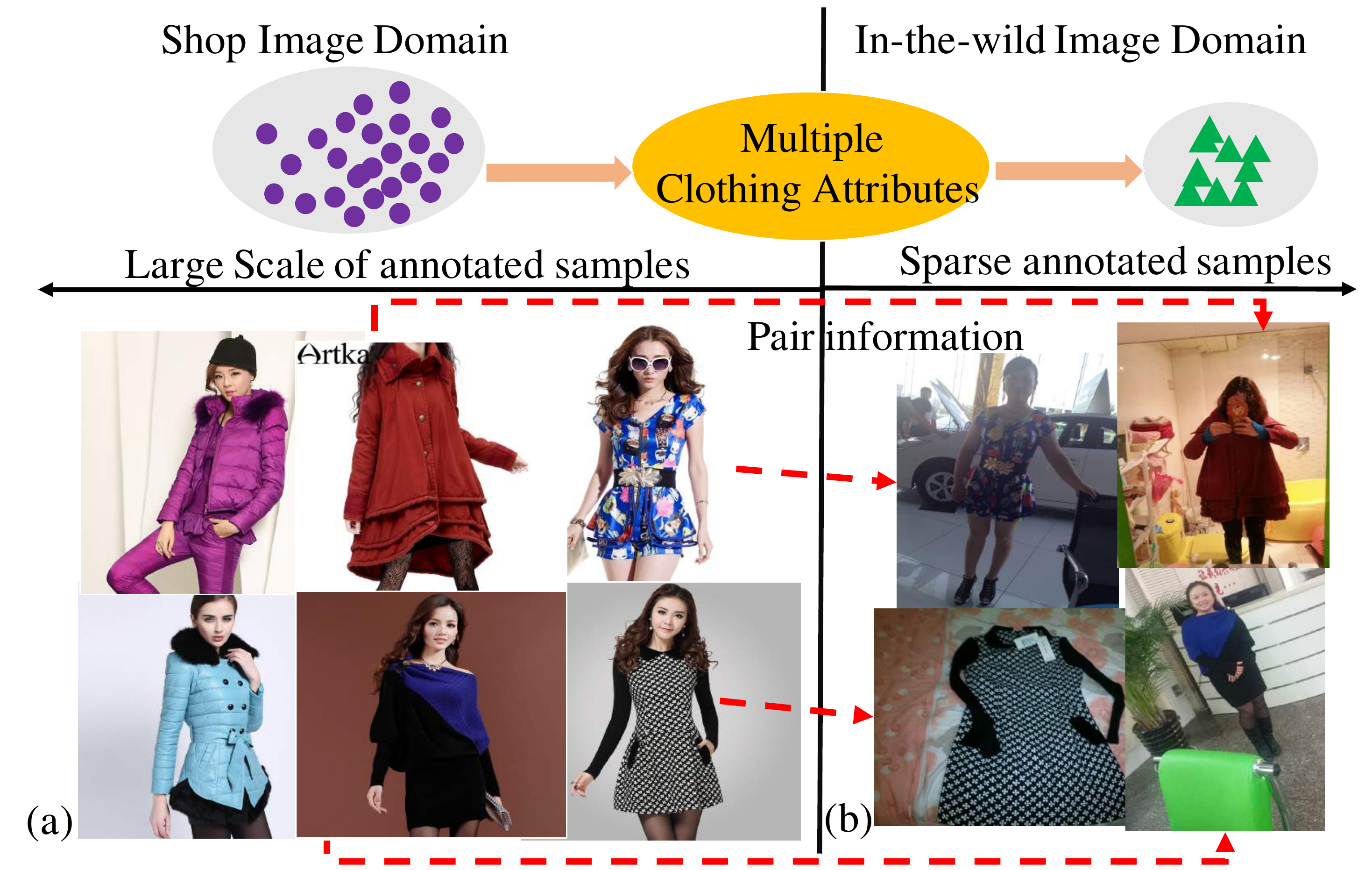}
	\caption{Clothing images of (a) professional models in shops
		and their corresponding instances (b) in-the-wild from
        the streets, with significant changes in appearance and
        background clutter.
	}
	\label{fig:problem}
\end{figure}

%
Automatic recognition of clothing attributes in images from the wild,
e.g. street views, has many applications from retail
shopping to internet search and visual surveillance \cite{gong2014person,feris2014attribute}.
%
However, clothing attribute recognition in-the-wild
is challenging due to poor lighting, cluttered scenes, unknown
viewpoint, and lacking image details (Figure \ref{fig:problem} (b)).
%
%
Deep learning exploits a large collection of imagery data from diverse
sources, and has been shown to be very
effective for image classification tasks \cite{simonyan2014very,sharif2014cnn,krizhevsky2012imagenet,bengio2013representation}.
However, training a deep model requires extensive labelled information
on the imagery data mostly generated by exhaustive manual annotation.
For clothing attribute, the available labelled image data size is
small and the number of fine-grained attribute categories is also
limited
\cite{feris2014attribute,vaquero2009attribute,layne2012person}.

To overcome the lack of manually labelled training data, web data mining
provides a solution
\cite{divvala2014learning,chen2015deep,huang2015cross}, from which a large
number of web images with their meta-data can be collected without
exhaustive manual labelling. For clothing, there is potentially a rich
source of web images and their meta-text provided by online shops, 
where these meta-text contain fine-grained clothing attributes \cite{chen2015deep,huang2015cross,liu2016deepfashion}
(Figure \ref{fig:visresult}).
A notable characteristics of these online shopping clothing images is
that they are well-posed by models captured against clean
background in good lighting. A key challenge is how to transfer
models trained using these clean shop images to recognise attributes in
images captured in-the-wild from the streets, known as the {\em domain
  drift} problem \cite{torralba2011unbiased,nguyen2015dash}, where
clean shop photos are considered as the {\em source data} from a
source domain whilst unconstrained images from the wild are the {\em
  target data} from a target domain.

This work proposes a novel deep learning approach to modelling
clothing attributes given a large sized {\em source data} and very small
sized {\em target data} for model training. The model solves the
challenging problem of transfer learning between the source and target
domains when both source and target training data are weakly labelled at the image level.
Our contributions are three-fold:
(1) We formulate a novel {\em Multi-Task Curriculum Transfer}
(MTCT) deep learning approach to modelling clothing attributes.
(2) In contrast to existing methods
\cite{chen2015deep,huang2015cross,liu2016deepfashion}, which are
limited in exploiting cross-domain data, the proposed MTCT deep
attribute model is characterised by
a multi-task joint learning deep network architecture
for capturing the underlying correlations between different attributes
with shared feature representations.
(3) We implement a novel curriculum transfer
deep learning strategy that aims to explore knowledge about
attributes for solving more effectively the highly non-convex
optimisation problem in model learning.
This curriculum transfer
learning strategy is motivated by cognitive studies
\cite{elman1993learning,rohde1999language,krueger2009flexible,bengio2009curriculum},
with a multi-staged learning principle focusing on a simple task first
before increasing the learning difficulty level, reminiscent to the
human learning strategy.
To our knowledge, this is the first attempt of formulating a
curriculum learning strategy for deep learning of attributes,
although there was an early study on language modelling
\cite{bengio2009curriculum}.
This is in strong contrast to the current popular end-to-end learning
strategy deployed in deep learning \cite{jia2014caffe,simonyan2014very}.
Our extensive comparative evaluation using the X-Domain benchmark
dataset \cite{chen2015deep} against three state-of-the-art deep
learning models for clothing attribute recognition, including
FashionNet \cite{liu2016deepfashion}, DARN \cite{huang2015cross} and DDAN
\cite{chen2015deep}, demonstrates a clear advantage of the proposed
MTCT deep attribute model. 
Moreover, we show
that MTCT also has a notable advantage over the
state-of-the-arts when the target domain training data size becomes small.

\section{Related Work}

\noindent {\bf Attributes. }
Visual attributes have been widely exploited in computer vision,
e.g. zero-shot learning \cite{lampert2009learning,fu2015transductive},
face analysis \cite{kumar2009attribute},
pedestrian description \cite{deng2014pedestrian,gallagher2008clothing},
person re-identification \cite{layne2012person},
visual search \cite{kovashka2012whittlesearch,siddiquie2011image,feris2014attribute}.
These studies typically pre-define a small set of attributes and
require expensive manual labelling of training data. 
Different datasets often do not have consistent labelling, limiting
their scalability.
%
Driven by the desire for large quantities of cheaply labelled images,
there are studies to explore web data sources for collecting large
scale imagery data that come with ``free'' corresponding meta-text
annotations
\cite{chen2013neil,divvala2014learning,rohrbach2010helps}.
However,
this poses a new problem in that the meta-data labels of these web data are less accurate nor consistent when compared with human manually labelled attributes.
%
%
%
Studies on clothing modelling have been focused extensively on
clothing segmentation against typically clean background
\cite{chen2012describing,kiapour2014hipster,liu2012street,yamaguchi2012parsing}.
There have also been efforts on shop-clothing image categorisation
and retrieval using traditional hand-crafted features (e.g. SIFT, HOG)
\cite{bossard2013apparel,chen2012describing,chen2012describing,fu2012efficient,wang2011clothes},
and more recently deep learning based features
\cite{huang2015cross,chen2015deep,kiapour2015buy,liu2016deepfashion}.
Given the costs of labelling therefore a lack of large scale clothing attribute
annotations from different sources (domains), cross-domain clothing
attribute learning is a challenging problem and largely under-studied
\cite{liu2016deepfashion,huang2015cross}.  








\vspace{0.1cm}
\noindent {\bf Deep Transfer Learning.}
%
Transfer learning for domain adaptation is a well studied area
\cite{gong2012geodesic,gopalan2014unsupervised,fernando2013unsupervised,shao2015transfer}.
More recently, deep learning models are shown to be more robust
than conventional models against domain changes, mainly due to
the high modelling capacity and the availability of large
scale labelled training data.
However, the domain drift problem remains unsolved
i.e. the performance of deep models still degrades in a new domain
\cite{hoffman2013one}.
A common approach to deep transfer learning is fine-tuning, using
target domain data, the higher layers (FC layers) of a pre-trained deep model
from the source data
\cite{girshick2014rich,sermanet2013overfeat}.
This assumes the availability of a large number of target training data,
which is mostly not the case.
A number of deep transfer learning models have been proposed for
generic image categorisation
\cite{hoffman2013one,hu2015deep,ding2016task,tzeng2015simultaneous,nguyen2015dash},
and more recently for fine-grained clothing attribute learning
\cite{chen2015deep,huang2015cross,liu2016deepfashion}.
Specifically,
%
Chen et al. \cite{chen2015deep} proposed a
Deep Domain Adaptation Network (DDAN) with two branches
by assigning one branch to a specific domain and then
introducing two cross-branch connected layers
that can enforce a feature distance between cross-domain images
according to their attribute relations.
A further extension of the DDAN model was also proposed by
Huang et al. \cite{huang2015cross}, which consists of a Dual
Attribute-aware Ranking Network (DARN) to additionally accommodate
image-level cross-domain correspondence as well as hierarchical
structural knowledge of attributes in each network branch.
More recently, Liu et al. \cite{liu2016deepfashion} introduced 
a FashionNet to model simultaneously both local attribute-level
and holistic image-level clothing
representations with a strong requirement on
manually labelled clothing landmarks, making it less
scalable than both DDAN and DARN networks.
Our new MTCT (Multi-Task Curriculum Transfer) network shares some
common characteristics with DDAN and DARN but also with a few important
differences and advantages:
Unlike DDAN, that only considers attribute labels, 
our method additionally models image-level cross-domain image pair
relations for more effective domain adaptation.
This is similar to DARN. However, whilst our domain transfer learning
exploits multi-task/attribute feature learning, DARN
only utilises shared common fully-connected feature representations
for all attributes (Figure \ref{fig:pipeline}(c)).
%
%
%
In contrast to the FashionNet, our MTCT net exploits cross-domain
attribute learning {\em without} the need for extensive clothing
landmarks, more scalable to wider applications.
Moreover, our MTCT network explores uniquely the curriculum transfer
learning strategy for more effective deep model learning. Our
extensive comparative evaluation validates the advantages of MTCT
over DDAN \cite{chen2015deep}, DARN \cite{huang2015cross}, and the
FashionNet \cite{liu2016deepfashion}. 
%








\section{Multi-Task Curriculum Transfer Network}
\label{sec:method}

\subsection{Problem  Definition}
\label{sec:problem}

To construct a deep model capable of recognising fine-grained clothing
attributes on images in-the-wild (target domain), we collect clothing
images and their meta-label as attributes
$\{z_i\}_{i=1}^{n_\text{attr}}$ (e.g. clothing category, collar style)
automatically from a range of online shopping web-sites, with a total
of $n_\text{attr}$ different attribute categories, each category $z_i$
having its respective value range $Z_i$.
Intrinsically, this is a {\em multi-label} recognition problem since
the $n_\text{attr}$ attribute categories co-exist in every clothing
image and may be assigned to different values. 

Suppose (1) we have a collection of $n_t$ {\em target} training images
$\{\bm{I}_i^t\}_{i=1}^{n_t}$ along with their attribute annotation vectors
$\{\bm{a}_i^t\}_{i=1}^{n_t}$,
and $\bm{a}_i^t=[a_{i,1}^t, \dots, a_{i,j}^t, \dots,a_{i,n_\text{attr}}^t]$
where $a_{i,j}^t$ refers to the $j$-th attribute value of target image $\bm{I}_i^t$;
there are also $n_s$ {\em source} training images
$\{\bm{I}_i^s\}_{i=1}^{n_s}$ with corresponding attribute vectors
$\{\bm{a}_i^s\}_{i=1}^{n_s}$;
$n_s >> n_t$, that is, the number of labelled source images is
much greater than that of labelled target images.
Moreover, (2) we have access to $n_\text{pw}$ {\em pair correspondences}
between target and source clothing images, e.g. selfie images taken by
shopping customers with known pairing to the online images of the same
clothes (Figure \ref{fig:problem}).
This cross-domain pair relation is useful in bridging
the large domain gap by transferring attribute knowledge encoded in
the source domain to the target domain with much less labelled data.
It is worth noting that these two types of supervised learning
lie at different levels: Most attributes are {\em localised} to image
regions, even though the location information is not
provided in the annotation.
Cross-domain pair labels are at the {\em holistic} image-level.
We consider this not only a multi-label learning problem -- joint
learning for mutually correlated attribute labels, but also a
{\em multi-task} transfer learning problem -- inter-dependently
learning the best individual attribute prediction given both local and
holistic cross-domain annotations.
%


\begin{figure*}[th]
\centering
\includegraphics[width=1\textwidth]{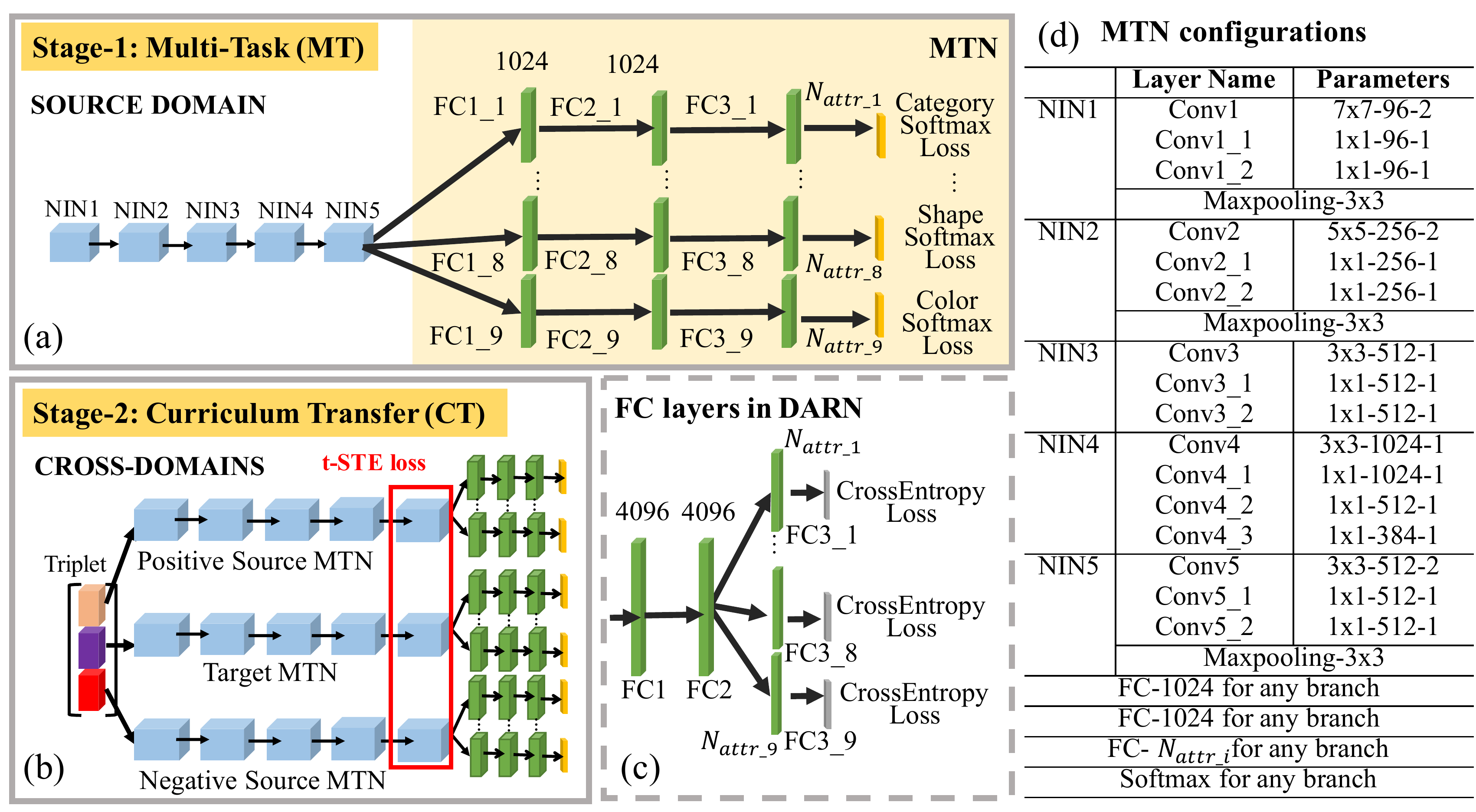}
\caption{
	(a)-(b): The MTCT network design, (c):
  The FC layers of DARN \cite{huang2015cross}, (d): MTN configuration
  details. }
	\vspace{-0.4cm}
\label{fig:pipeline}
\end{figure*}

\subsection{Network Architecture Design}
\label{sec:our_network}

Our MTCT deep model has two components:
(I) multi-task deep learning,
(II) curriculum transfer deep learning. 

\vspace{0.1cm}
\noindent {\bf (I) Multi-Task Deep Learning. }
Clothing attributes co-occur selectively and to explore this inherent
constraint for more reliable attribute prediction, we wish to model
multi-attribute correlations by formulating a {\em Multi-Task
  Network} (MTN).
This implements the multi-task learning principle
\cite{evgeniou2004regularized,ando2005framework} in a deep model.
Although sharing a similar spirit of multi-task {\em regression}
networks for face modelling \cite{zhang2014facial, yim2015rotating},
in this MTN model we learn a multi-task {\em discriminative} network for
clothing modelling.
%
%
Compared to independent attribute modelling,
such multi-task learning also involves
a smaller number of to-be-learned model parameters and thus
with a lower model overfitting risk towards the given training data,
beyond modelling mutual relations among different types of attributes
and their common representations.
%

Specifically, the MTN consists of
five stacked Network-In-Network (NIN) convolutional (conv) units \cite{lin2013network} and
$n_\text{attr}$ parallel branches, with each branch representing
a three layers of Fully-Connected (FC) sub-network for modelling
one of the $n_\text{attr}$ attributes respectively
(Figure \ref{fig:pipeline}(a,d)).
The neuron number in the output-layer of $i$-th branch is $|Z_i|$,
i.e. the number of corresponding all possible attribute values
$a_i$.
%
For model training,
we utilise the Softmax loss function
within any branch to model mutually exclusive relations among the
attribute values for each attribute category by firstly
predicting the $j$-th attribute posterior probability of image $\bm{I}_i$
over the ground truth $a_{i,j}$:
\begin{equation}
{p}(y_{i,j} = a_{i,j} | \bm{x}_{i,j}) = \frac{\exp(\bm{W}_{j}^{\top} \bm{x}_{i,j})} {\sum_{k=1}^{|Z_{j}|} \exp(\bm{W}_{k}^{\top} \bm{x}_{i,j})}
\end{equation}
where $\bm{x}_{i,j}$ refers to the feature vector for $j$-th attribute,
and $\bm{W}_k$ to the corresponding prediction function parameter,
then computing
the overall loss on a batch of $n_\text{bs}$ images as
the average additive summation of attribute-level loss with equal weight: 
  \begin{equation}
  l_\text{sm} = - \frac{1}{n_\text{bs}}\sum_{i=1}^{n_\text{bs}}  \sum_{j=1}^{n_\text{attr}} \log \Big(p(y_{i,j}=a_{i,j}|\bm{x}_{i,j}) \Big)
 \label{eq:loss}
\end{equation}
This design above allows to jointly learn both attribute-generic (by all shared conv layers)
and attribute-specific (by individual FC layer branches) discriminative features.
In this context, each branch corresponds to a specific learning task responsible for
the assigned attribute modelling.

The proposed MTN is similar to the DARN model \cite{huang2015cross}
but with a crucial difference, that is,  
the attribute-specific branch in DARN
contains only the last FC$_3$ layer
which serves as an attribute prediction function (Figure \ref{fig:pipeline} (c)),
therefore no attribute-specific representation learning in DARN
and our experiments show this is less effective in learning
discriminative features as compared to the proposed MTN, where
FC$_{1,2}$ layers are explicitly allocated for this purpose in each branch.
As all clothing attributes are jointly modelled in DARN,
we refer it as {\em Joint Attribute Convolutional Neural Network}
(JAN) in the experimental evaluation reported in Section 4.

Learning the MTN model requires a large amount of training
data\footnote{Each MTN has $19$ conv, $3$ max-pooling
  and $27$ ($9\times3$) FC layers. The MTCT model has in total
  $79.4$ million parameters (57 million required fine-tuning), In comparison,
  DARN \cite{huang2015cross} and FashionNet \cite{liu2016deepfashion}
  have 73 million and 135 million parameters respectively (all
  required fine-tuning).}, 
whilst we usually only have very limited labelled target images.
To overcome this problem, we want to explore richer source domain labelling
information through cross-domain transfer learning.


\vspace{0.1cm}
\noindent {\bf (II) Curriculum Transfer Learning. }
To transfer source annotation knowledge to
sparsely labelled target domain,
we formulate a {\em Curriculum Transfer} (CT) learning strategy for
deep learning.
This is motivated by cognitive studies that suggest
a better learning strategy adopted by human and animals is to start with
learning easier tasks before gradually increasing the
difficulties of the tasks, rather than to blindly learn
randomly organised tasks
\cite{elman1993learning,rohde1999language,krueger2009flexible}.

In our cross-domain clothing attribute learning context,
source and target domain data present naturally this easy-hard
knowledge distribution:
Source images taken from professional
models are much easier to learn than target images captured
in-the-wild (Figure \ref{fig:problem});
Moreover, the difference in holistic (cross-domain pairing) and local
annotations (source domain attributes)
exhibits distinct degrees of learning complexity --
attributes are localised thus specific
whilst pair correspondences are holistic therefore abstract,
with the latter has greater variations than the former.

Given these observations above, we propose a {\em two-stage curriculum
  transfer} (CT) learning strategy
tailored for clothing attribute modelling:
(1) Stage-1: Learning with clean (easier) source images and their
attribute labels only;
(2) Stage-2: Learning to capture harder cross-domain knowledge
by embedding cross-domain image pair information, and simultaneously
appending harder target images into the model training process.
As such, attribute labels of non-paired source images can also be
exploited in addition to the cross-domain paired images.
%
%
We instantiate this CT strategy by formulating a MTN based deep architecture.

\vspace{0.1cm}
\noindent {\em CT Learning Stage-1: Easy Model Learning and Transfer}.
The main purpose of this first CT learning stage is to
extract well-defined attribute features from large quantities of
easy-to-learn source images and then directly transfer to the target domain.
The intuition is that, the high non-convex model
optimisation problem we need to address can be made not only simpler
but also benefiting from better identified local minima for subsequent
incremental learning if the starting sub-tasks are restricted to less
hard tasks \cite{bengio2009curriculum}.  
This principle is consistent with the notion of {\em adaptive value} of
  starting in developmental psychology \cite{elman1993learning}.
This direct feature transfer from source to target domain
exploits the characteristics of deep learning features being capable
of capturing hierarchical information and independent of
the training data, particularly from the lower layers
\cite{yosinski2014transferable,jia2014caffe,sharif2014cnn,sermanet2013overfeat}.

%


The easy-stage (Stage-1) of our CT learning strategy constructs a
source and a target MTN model as follows:
(1) Following common practice \cite{simonyan2014very},
we pre-train all NIN layers of a source MTN
using the training data of ImageNet-1K \cite{russakovsky2015imagenet}
for obtaining a good parameter initialisation.
(2) We train the whole MTN on source images with their attribute labels.
(3) We create a target MTN by sharing all parameters from the source MTN.
In this way, the attribute information is transferred from source
to target domain.
%
%

\vspace{0.1cm}
\noindent {\em CT Learning Stage-2: Hard Model Learning and Transfer}.
We build on Stage-1 to
transfer {\em harder} cross-domain pair relational knowledge
and perform {\em incremental} learning on harder target data.
This is achieved by constructing a three-stream MTN (3MTN) architecture
consisting of two identical copies of the source MTN network and the
target MTN obtained from Stage-1 (Figure \ref{fig:pipeline}(b)),
taking as input
cross-domain image triplets in the form of
``\{{\em Target} $\bm{I}_\text{t}$, {\em Positive Source} $\bm{I}_\text{ps}$,
{\em Negative Source} $\bm{I}_\text{ns}$\}''
where {\em Target} $\bm{I}_\text{t}$ and
{\em Positive Source} $\bm{I}_\text{ps}$ are of the same clothing
(obtained from the cross-domain pairing labelling),
whereas {\em Target} $\bm{I}_\text{t}$ and {\em Negative Source}
$\bm{I}_\text{ns}$ are of different clothing items.
We require that feature similarity value between {\em Target}
and {\em Positive Source} is greater than that between {\em Target}
and {\em Negative Source} simultaneously.
To this end, we consider the {\em learning-to-rank} approach to model optimisation and
exploit the {\em t-distribution Stochastic Triplet Embedding} (t-STE)
loss function due to its strength in discovering underlying data
structure \cite{van2012stochastic}:
\begin{align}
& l_\text{t-STE} \; = \;  \sum_{ \{\bm{I}_\text{t},\bm{I}_\text{ps},\bm{I}_\text{ns}\} \in T}
\\ \nonumber
& \log \frac{(1+\frac{{\|f_\text{t}(\bm{I}_\text{t})-f_\text{s}(\bm{I}_\text{ps})\|}^{2}}{\alpha})^\beta}
{(1+\frac{{\|f_\text{t}(\bm{I}_\text{t})-f_\text{s}(\bm{I}_\text{ps})\|}^{2}}{\alpha})^\beta
	+ (1+\frac{{\|f_\text{t}(\bm{I}_\text{t})-f_\text{s}(\bm{I}_\text{ns})\|}^{2}}{\alpha})^\beta}
\label{eq:ste}
\end{align}
where
$\alpha$ denotes the freedom degree of the Student kernel;
$\beta = {-\frac{(1+\alpha)}{2}}$;
$f_\text{t}(\cdot)$ and $f_\text{s}(\cdot)$ refer to the feature extraction function
for the target and source MTN respectively, that is,
the vectorised feature maps of the conv$5$ layer used as the sample feature
in each stream (Figure \ref{fig:pipeline}(b)).
%

Concurrently, we learn all FC layers of each attribute-specific branch
in the {\em Target} stream with the Softmax loss
for obtaining the final attribute recognition model
(Figure \ref{fig:pipeline}(b)).
In practice, we found that fine-tuning FC layers in the source stream
helps due to the mutual benefits between the two domains.
As a result, all layers are {\em frozen} except conv$5$ of the {\em
  Target} MTN stream and all FC layers of both streams during the CT
Stage-2 learning. \\

\noindent {\bf Clothing Attribute Recognition}:
\label{sec:attrRecg}
Our learning aim is to obtain an optimised target MTN model for
attribute recognition in-the-wild.
This is achieved during model training by
learning a source MTN for extracting and transferring
localised attribute information, followed by optimising a 3MTN
for transferring cross-domain pairing knowledge
and adapting the target MTN to data from the wild.
During model deployment, we solely utilise the target MTN
for fine-grained clothing attribute recognition on unconstrained images.
In the next section, we shall demonstrate the effectiveness of the
proposed model when compared against the state-of-the-arts.

\section{Experiments}
\label{sec:Exp}

\subsection{Dataset and Evaluation Protocol}
\label{sec:dataset}
We utilised the Cross-Domain (X-Domain) clothing attribute dataset \cite{chen2015deep}
for our comparative evaluations\footnote{In our experiments, we
  collected 100\% shop images of the X-Domain dataset, but only 69\%
  of the cross-domain pairing images were available from the X-Domain
  URLs given by \cite{chen2015deep} due to commercial copyrights.}.
Specifically, this X-Domain dataset contains two different image
source domains: (1) The {\em shop} domain, online stores
such as Amazon.com and TMall.com;
(2) The {\em street} domain where consumer images
are available.

Specifically,
there are $245,467$ shop images each associated with web meta-data including $\leq$ $9$ attribute/value pairs.
These nine fine-grained clothing attributes are:
{\em category},
{\em button},
{\em colour},
{\em length},
{\em pattern},
{\em shape},
{\em collar},
{\em sleeve-length} (slv-len),
{\em sleeve-shape} (slv-shp).
There may be varying numbers of optional values for different attributes,
ranging from $6$ (slv-len) to $55$ (colour) and
a total of $178$ distinct values over all attributes.
Therefore, these clothing attributes are rather {\em fine-grained},
possibly with subtle visual appearance dissimilarity between different
attribute values, e.g. Woollen-Coat {\em versus} Cotton-Coat.
Note that these attribute data were {\em webly annotated} at the {\em
  image-level} and thus {\em weakly-supervised} with no specified
attribute location.

We also have $46,769$ street images
from customer reviews of a proportion of shop image webpages.
Among these $46,769$ in-the-wild images, there are $14,186$ cross-domain
pairing with the shop images. The remaining
$231,281$ $(245,467-14,186)$ shop images are non-paired.
%
%
In our evaluations, we consider the shop and street domains as the
{\em source} and {\em target} domains, respectively.

\begin{table*}[th] \footnotesize
	\centering
	\setlength{\tabcolsep}{0.2cm}
	\caption{
		Comparing state-of-the-art clothing attribute recognition methods.
	}
	\label{tab:art}
	\begin{tabular}{c||c|c|c|c|c|c|c|c|c||c||c|c}
		\hline
		Method & Category & Button & Colour & Length & Pattern & Shape & Collar & Slv-Len & Slv-Shp & {mAP$^\text{cls}$}
		& mP$^\text{ins}$ & mR$^\text{ins}$
		\\ \hline \hline

		DDAN \cite{chen2015deep}
		& 12.56& 24.13 & 20.72 & 35.91& 61.67 & 47.14 & 31.17 & 80.63 & 73.96&  {43.10}
		&45.41&52.20
		\\ \hline
		
		DARN \cite{huang2015cross}
		& 52.55 &  37.48 & 58.24 & 51.49 & 67.53 & 47.70 & 47.77 & 82.04 & 73.72& {57.61}
		&57.79&67.29
		\\ \hline
		
		FashionNet \cite{liu2016deepfashion}
		& 55.85 & 39.52 & 60.33 & 53.08 & 68.65 & 49.79 & 52.17 & 83.79 & 75.34& 59.84 & 59.97 &  69.74
		\\ \hline
		{\bf MTCT }
		& {\bf 65.96} & {\bf 43.57} & {\bf 66.86} & {\bf 58.27} & {\bf 70.55} & {\bf 51.40} & {\bf 58.97} & {\bf 86.05} & {\bf 77.54} & {\bf 64.35}
		& {\bf 64.97} & {\bf 75.66}
		\\ \hline
	\end{tabular}
	\vspace{-0.4cm}
\end{table*}

\vspace{0.1cm}
\noindent {\bf Evaluation Protocol. }
On our copy of the X-Domain dataset, we performed the following data partition
for cross-domain attribute recognition evaluation.
For the shop domain, we randomly selected $165,467$ images as training data
and the remaining $80,000$ as test images. For the street domain,
$36,769$ were randomly selected for training and $10,000$ for
test. 

For quantitative evaluation, we adopted
both {\em per-class} (i.e. per-attribute) and {\em per-instance} (i.e. per-image) based metrics.
For the former, we used
Average Precision (AP$^\text{cls}$) for each attribute class and
mean Average Precision (mAP$^\text{cls}$) over all classes \cite{chen2015deep}.
For the latter,
we first computed per-image attribute Precision and Recall,
then averaged both over all images
to obtain mean Precision (mP$^\text{ins}$)
and Recall (mR$^\text{ins}$) \cite{liu2016deepfashion}.



\subsection{Implementational Considerations}
\label{sec:impl}

\noindent {\bf Clothing Detection.} As input images are not accurately
cropped, clothing detection is necessary for reducing the negative impact of
background clutter. We performed clothing detection by a customised
Faster R-CNN model \cite{ren2015faster}.
Specifically,
we first trained our detector on PASCAL VOC2007 training data \cite{everingham2010pascal}
followed by fine-tuning on an assembled clothing dataset
consisting of
$8,000$ street/shop photos (with box annotation available) from \cite{kiapour2015buy}
and $4,000$ fashion images (with boxes generated from available pixel-level labels) from \cite{liang2015deep}.

\vspace{0.1cm}
\noindent {\bf Parameter Settings.} For training the MTCT, the momentum was set at
0.9 and weight-decay at 0.0005, same as in NIN \cite{lin2013network} and AlexNet \cite{krizhevsky2012imagenet}. 
	The batchsize was set at 256 limited by the GPU memory
        size. The learning rates were set empirically, by the
        training loss change, at 0.001 for pre-training on the ImageNet, and 0.0001 for fine-tuning on
        the source/target clothing data. For training the other
        compared models, we used the same parameter settings given by the authors, otherwise same as for MTCT.


\subsection{Evaluation Choices}
We compared our MTCT deep attribute model with 3 state-of-the-art
models and 4 different variants of our model design:
{\bf (1)} {\em Deep Domain Adaptation Network} (DDAN) \cite{chen2015deep}:
A cross-domain attribute recognition model
capable of learning domain invariant features by particularly
aligning middle level representations of two domains during the training stage.
{\bf (2)} {\em Dual Attribute-aware Ranking Network} (DARN) \cite{huang2015cross}:
A domain adaptation deep model that is trained and optimised
with both attribute annotations and cross-domain pair correspondences
in an end-to-end learning manner.
{\bf (3)} {\em FashionNet} \cite{liu2016deepfashion}: A very recent
clothing analysis model specially designed for multiple recognition
tasks such as attribute and landmark detection. We implemented this
model excluding landmark detection branch since landmark labels are
not available in this real-world X-Domain dataset.
We also compared MTCT against four different MTN based
models to evaluate the role of the individual components in the MTCT
model design.
These are: {\bf (4)}
{\em No Adaptation} (NoAdpt):
We train a given network using the labelled source data and directly
deploy it for the target test data.
This simple scheme has shown power and superiorities in many applications \cite{sharif2014cnn,sermanet2013overfeat,jia2014caffe}
due to the great generality of deep features by benefiting from large scale diverse training data.
{\bf (5)}
{\em JAN} (NoAdpt): we set DARN without adaptation as baseline, i.e. training JAN in the source domain and then directly testing it on the target domain.
{\bf (6)} {\em United Domains} (UD):
We train a given model on the union of source and target training data.
Compared with NoAdpt, more data are exploited for model optimisation so that
the feature generality may be further improved.
{\bf (7)} {\em Fine-Tune based Transfer} (FTT):
We first train a given model on the source training data,
then fine-tune the fully-connected layers on the target data.
This is the vanilla transfer learning method commonly
adopted in the literature \cite{yosinski2014transferable}.
Finally, we have the {\bf (8)} {\em Multi-Task Curriculum Transfer}
(MTCT): Our full model
exploiting both multi-task and curriculum transfer learning.
For fair comparison, all methods have access to the same training data,
learned with their designed optimisation algorithms,
and evaluated on the same test set.

%
%
%
%
%

\begin{table*}[th] \footnotesize
	\centering
	\setlength{\tabcolsep}{0.2cm}
	\caption{
		Evaluating the effects of multi-task and curriculum
                transfer learning in MTCT.
	}
	\label{tab:mtn}
	\begin{tabular}{c||c|c|c|c|c|c|c|c|c||c||c|c}
		\hline
		Method & Category & Button & Colour & Length & Pattern & Shape & Collar & Slv-Len & Slv-Shp & {mAP$^\text{cls}$}
		& mP$^\text{ins}$ & mR$^\text{ins}$
		\\ \hline \hline
		JAN(NoAdpt) \cite{huang2015cross}
		& 34.08 & {35.87} & 43.08 & 44.21 & 63.76 & 43.40 & 40.50 & 78.13 & 71.08 & {50.46}
		&50.39&58.40 
		\\ \hline
		
		MTN(NoAdpt) 
		& {35.77} & {33.77} & {44.13} & {44.76} & {65.26} & {45.75} & {40.85} & {79.76} & {72.40} & {51.38}
		& {51.82} & {60.00} 
		\\ \hline
		MTN(UD)
		& 54.10&40.65  &57.88 &51.35&67.80&49.79&49.09&83.61&74.60& 58.76&60.16 &70.00   
		\\ \hline
		MTN(FTT)
		&61.92& 42.65 & 65.43 & 55.16  & 70.06 & 49.00  & 50.55 & 85.54 & 76.04 & {61.82}
		&62.53&72.76
		\\ \hline
		{\bf MTCT }
		& {\bf 65.96} & {\bf 43.57} & {\bf 66.86} & {\bf 58.27} & {\bf 70.55} & {\bf 51.40} & {\bf 58.97} & {\bf 86.05} & {\bf 77.54} & {\bf 64.35}
		& {\bf 64.97} & {\bf 75.66}
		\\ \hline
	\end{tabular}
	\vspace{-0.3cm}
\end{table*}

\begin{figure*}[th]
	\centering
	\includegraphics[width=1\textwidth]{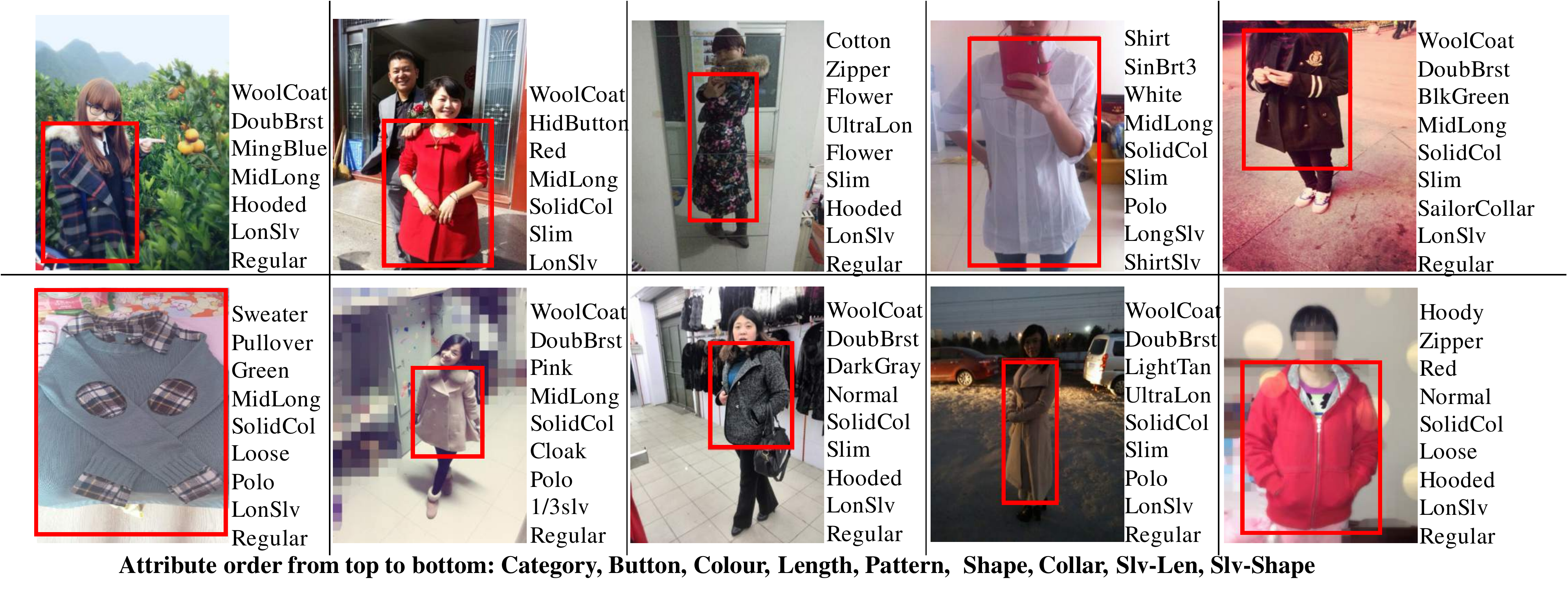}
	\vskip -0.2cm
	\caption{
		A qualitative evaluation of our proposed MTCT method
		on unconstrained consumer images. 
	}
	\vspace{-0.4cm}
	\label{fig:visresult}
\end{figure*}

\begin{figure}[t]
	\centering
	\includegraphics[width=0.9\linewidth]{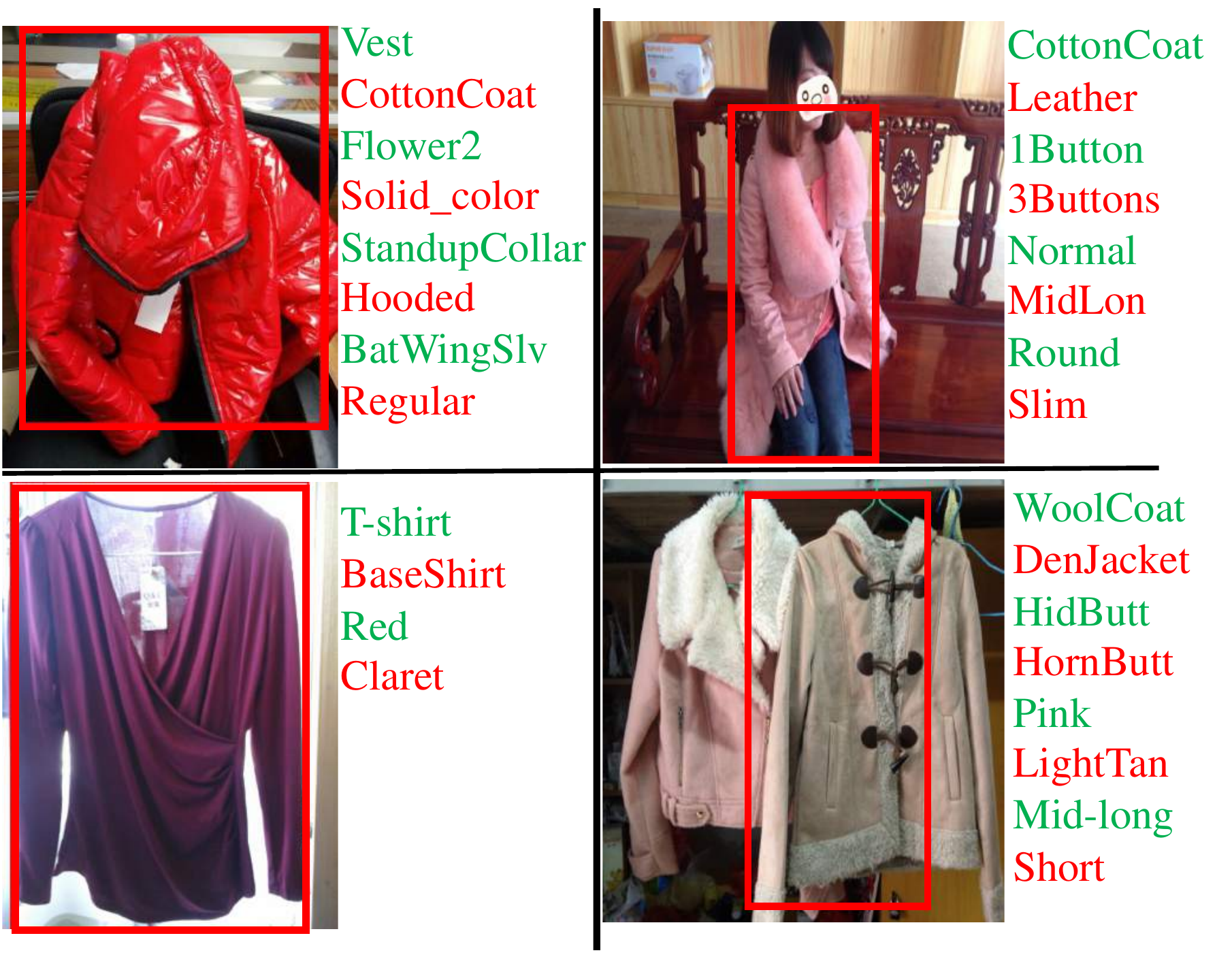}
	\caption{
		Failure cases: Incorrect attribute predictions (green)
                are shown against the corresponding ground truth (red) underneath.
		}
	\vspace{-0.5cm}
	\label{fig:fail}
\end{figure}


\subsection{Comparison to the State-Of-The-Art}
\label{sec:compareStateArt}
We evaluated comparatively MTCT model performance on clothing
attribute recognition.
The comparative results with state-of-the-arts are presented in Table \ref{tab:art}.
It is evident that the proposed MTCT achieves the best results 
under all evaluation metrics, e.g.
outperforming the best alternative FashionNet
by $4.51\%$ in mAP$^\text{cls}$.
This suggests the superiority of our method in extracting and transferring source annotation information
into the sparsely labelled and challenging target domain.
More specifically, we can draw the following observations.
%
%
Firstly, DDAN is the worst performer among all competitors,
mainly because this model is less effective in mining rich non-paired source images,
e.g. optimised with cross-domain paired images
whilst most non-paired ones are not selected for model learning.
By jointly modelling data from
both domains with additional pair relations,
DARN is able to extract and transfer more source information.
However, it also suffers from the same problem above as DDAN.
FashionNet surpasses DARN by learning the union of both domains and
exploiting 
a more powerful basis architecture 
VGG16 \cite{simonyan2014very} which is stronger than both the AlexNet
\cite{krizhevsky2012imagenet} used by DDAN and NIN
\cite{lin2013network} used by DARN.
Despite that, FashionNet is still inferior to the NIN based MTCT model
due to the former's higher model overfitting risk caused by much more
parameters required learning ($135$ million of FashionNet vs. $57$
million of MTCT parameters required fine-tuning in model learning) and
the ignorance of domain discrepancy in learning strategy (end-to-end
vs. curriculum staged learning).

\subsection{Effects of Multi-Task and Transfer Learning}
\label{sec:evalTransfer}
We evaluated the effectiveness of the multi-task and curriculum
transfer learning components in the MTCT model (Table~\ref{tab:mtn}).
%
By explicitly learning individual attribute representations,
the MTN(NoAdpt) without curriculum transfer improves model
generalisation over JAN(NoAdpt) (DARN \cite{huang2015cross} without
transfer learning). This demonstrates the benefit of multi-task learning.
When additional $36,769$ labelled target images (vs. $245,467$ source images)
were utilised, MTN(UD) improves further attribute recognition accuracy over
MTN(NoAdpt). 
This supports the general observation that learning from target domain
data is beneficial when there is a large discrepancy between the
source and target domains.
%
Given a vanilla fine-tuning transfer learning MTN(FTT),
%
model performance is further boosted, 
which confirms similar findings elsewhere \cite{bengio2012deep,bengio2011deep}.
MTN(FTT) is a special case of Curriculum Learning in
that the initialisation by source data is an easier learning task whilst
fine-tuning on target data is a harder task, but {\em without}
learning cross-domain pairing information.

Our MTCT model is a fusion of DARN and vanilla transfer learning
MTN(FTT) with unique advantages over both:
(1) Similar to DARN,
MTCT exploits cross-domain {\em pairing} information in model optimisation
but with a lot less parameters ($57$ million vs. $73$ million of
DARN); also different from DARN in that MTCT adopts curriculum learning rather
than end-to-end, first learning from easier attribute labels (source) then from
harder pairing relations.
(2) Similar to FTT, MTCT is optimised
in a staged process, first learning easier source data then harder
target data, plus harder still pairing data.
Therefore, MTCT model explores two curriculum learning criteria,
one on training data selection and another on supervision label difficulties.
Qualitative evaluation is shown in Figure \ref{fig:visresult}, where
success cases showing the robustness of MTCT against cluttered
background and complex viewing conditions. Figure \ref{fig:fail} shows
some failure cases under extreme poses and very challenging
background ambiguities.


%
%
%
%
%


\begin{figure}
	\centering
	\includegraphics[width=0.7\linewidth]{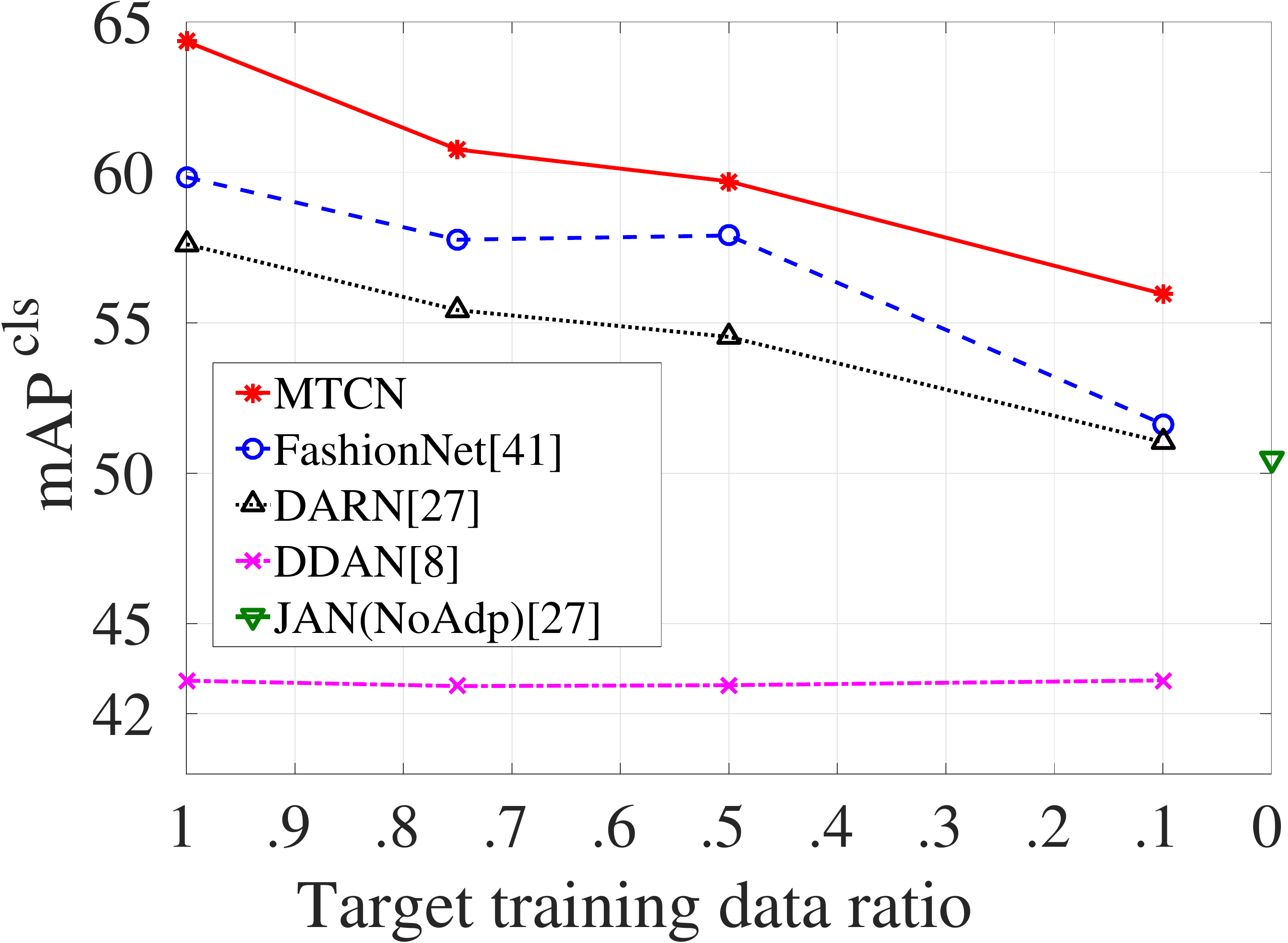}
	\vskip -0.1cm
	\caption{Model robustness vs. target training
          data size. }
	\label{fig:sparse}
	\vspace{-0.4cm}
\end{figure}

\begin{table*}[th]\footnotesize
	\centering
	\caption{Comparing curriculum vs. end-to-end transfer learning
          using the MTN network.
	} 
	\setlength{\tabcolsep}{0.2cm}
	\label{tab:cl}
	\begin{tabular}{c||c|c|c|c|c|c|c|c|c||c||c|c}
		\hline
		Method & Category & Button & Colour & Length & Pattern & Shape & Collar & Slv-Len & Slv-Shp & mAP$^\text{cls}$ & mP$^\text{ins}$ & mR$^\text{ins}$
		\\ \hline \hline 
		End-to-End
		&{61.11} & 41.39& 63.66& 56.29& 70.02& 51.39&55.45 & 84.69& 76.69 & {62.30}
		& 63.00 &73.37
		\\ \cline{1-13}
		{\bf Curriculum } & {\bf 65.96} & {\bf 43.57} & {\bf 66.86} & {\bf 58.27} & {\bf 70.55} & {\bf 51.40} & {\bf 58.97} & {\bf 86.05} & {\bf 77.54} & {\bf 64.35}
		&{\bf 64.97} & {\bf 75.66}
		\\\hline 
	\end{tabular}
	\vspace{-0.2cm}
\end{table*}

\begin{table*} 
	\footnotesize
	\centering
	\caption{Comparing different loss functions in the MTCT network.
		}
	\setlength{\tabcolsep}{0.2cm}
	\label{tab:loss}
	\begin{tabular}{c||c|c|c|c|c|c|c|c|c||c||c|c}
		\hline 
		Method & Category & Button & Colour & Length & Pattern & Shape & Collar & Slv-Len & Slv-Shp & mAP$^\text{cls}$ & mP$^\text{ins}$ & mR$^\text{ins}$ 
		\\ \hline \hline 
		triplet ranking \cite{huang2015cross} & 63.57&42.01&63.80&56.16&69.37&50.58&57.03&85.24&75.60& 62.60&63.45 &73.83
		\\ \hline \hline 
		{\bf t-STE \cite{van2012stochastic}} &{\bf 65.96}&{\bf 43.57}&{\bf 66.86}&{\bf 58.27}&{\bf 70.55}&{\bf 51.40}&{\bf 58.97}&{\bf 86.05}&{\bf 77.54}& {\bf 64.35}&{\bf 64.97}&{\bf 75.66}
		\\\hline 
	\end{tabular}
\vspace{-0.3cm} 
\end{table*}

\subsection{Effects of Cross-Domain Training Data Size}
\label{sec:limited_training_data}

We evaluated the robustness of different models
against target training data size variation.
For this evaluation, we reduced the number of target training images
to $\{75\%, 50\%, 10\%\}$ of the full training set and show respective
results in Figure \ref{fig:sparse}.
It is evident that the proposed MTCT outperforms
all competitors over different sparseness ratios.
This demonstrates the advantages and scalability of our approach over
alternatives models.
Specifically, JAN(NoAdpt) utilises no target data so remains at just
above $50\%$ mAP at a constant
(the green dot on the right hand side above $0\%$).
DDAN stays at a low $42-45\%$ with little change, suggesting that
transfer learning is difficult without exploiting cross-domain pair
relations.
As expected, the three models which have benefited from cross-domain pairing
information all degrade with fewer training data available.
Importantly, the MTCT model surpasses significantly other two models
with $8.4\%$ relative improvement over the FashionNet, given only
$3,676$ labelled target images.

\subsection{Further Analysis}
\label{sec:ablation}

\begin{figure}[h]
   \begin{center}
	\includegraphics[width=0.8\linewidth]{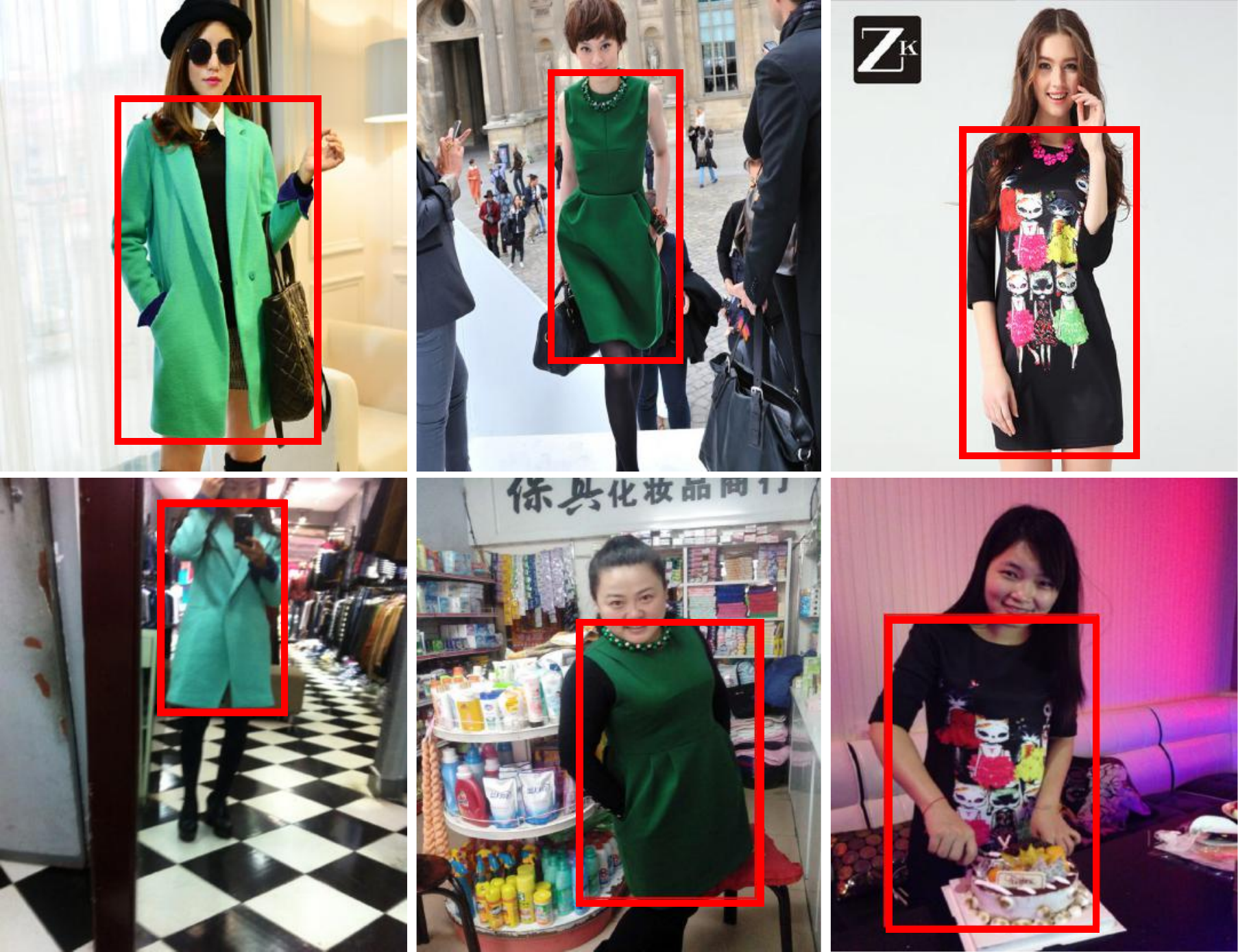}
   \end{center}
    \vskip -0.3cm
	\caption{
		Examples of attribute based automatic clothing
                detection and matching in-the-wild (bottom) given
                clean shop/model samples (top), or vice versa. Each pair of
                images in each column is of the same clothing item matched
                from different domains likely on different people.
	}
	\vspace{-0.5cm}
	\label{fig:det}
\end{figure}

\vspace{0.1cm}
\noindent \textbf{(1) Automatic Clothing Detection.} We evaluated the
performance of our customised Faster R-CNN 
clothing detector. For this evaluation, we manually labelled
clothing boxes on $400$ images from X-Domain,
including $200$ shop and $200$ consumer images.
We set the correct detection Intersection over Union (IoU) threshold to $0.6$.
Our detector achieves $90.8\%$ recall on shop images and
$71.2\%$ on in-the-wild images.
This provides a more realistic testing platform for attribute recognition.
Qualitative examples of clothing attribute detection and recognition,
failure cases, and cross-domain clothing matching by attributes (red
boxes) are shown in Figures 
\ref{fig:visresult}, \ref{fig:fail} and \ref{fig:det} respectively.

\vspace{0.1cm}
\noindent \textbf{(2) Triplet Ranking vs. t-STE loss.} Table \ref{tab:loss} compares the
performance of t-STE loss \cite{van2012stochastic} against the common
triplet ranking loss \cite{huang2015cross} in our MTCT network,
showing that the t-STE loss function yields mAP $1.75\%$ performance advantage
over the commonly used triplet ranking loss.

\vspace{0.1cm}
\noindent \textbf{(3) Curriculum vs. End-to-End Transfer Learning. }
We evaluated the effectiveness of our CT method
by comparing it with the popular End-to-End counterpart, both using our MTN architecture.
This End-to-End baseline can be considered as
an improved DARN model, i.e. replacing the JAN component of DARN with MTN.
The results are shown in Table \ref{tab:cl}.
It is clear that the proposed curriculum transfer is superior to
end-to-end transfer learning, an improvement of $2.05\%$ in
mAP$^\text{cls}$, suggesting that staged learning
can better regularise deep model optimisation
towards more discriminative local minima in the parameter space.


\section{Conclusion}
In this work, we formulated a Multi-Task Curriculum Transfer (MTCT) deep
learning method for modelling fine-grained clothing attributes. We
demonstrated its effectiveness in attribute recognition given
unconstrained images taken from-the-wild (street views).
This MTCT model (with $79.4$ million parameters) outperforms the
state-of-the-art FashionNet (with $135$ million parameters) by
$4.51\%$ in mAP$^\text{cls}$ on the X-Domain benchmark.
The proposed MTCT model is designed to optimise information transfer
learning given large quantities of labelled information in a clean
source domain and small sized labelled data in a noisy target
domain in-the-wild. Specifically, MTCT exploits both a multi-task
attribute learning deep network (MTN) and a staged curriculum learning
strategy to maximise model learning.
Moreover, we show the advantages of the MTCT over alternative models given
decreased sizes of labelled target domain data, surpassing the
FashionNet in performance on the X-Domain benchmark by $\sim$$8\%$
when only $<$$4,000$ target training images are available.

{\small
\bibliographystyle{ieee}
\bibliography{clothingAttr}
}

\end{document}